\documentclass{article}

\usepackage[nonatbib, final]{neurips_data_2024}

\usepackage[utf8]{inputenc} 
\usepackage[T1]{fontenc}    
\usepackage{hyperref}       
\usepackage{url}            
\usepackage{booktabs}       
\usepackage{amsfonts}       
\usepackage{nicefrac}       
\usepackage{microtype}      
\usepackage{xcolor}         
\usepackage{graphicx}

\usepackage{biblatex} 
\title{PLT-D3: A High-fidelity Dynamic Driving Simulation Dataset for Stereo Depth and Scene Flow}

\author{%
  Joshua~Tokarsky \\
  Pilotier\\
  Toronto, ON, Canada \\
  \texttt{josh@pilotier.com} \\
  \And
  Ibrahim~Abdulhafiz \\ 
  Department of Mech. \& Mechatron. Eng.\\
  University of Waterloo \\
  Waterloo, ON, Canada \\
  \texttt{i6abdulh@uwaterloo.ca} \\
  \AND
  Satya Ayyalasomayajula\\
  Pilotier\\
  Toronto, ON, Canada \\
  \texttt{satya@pilotier.com} \\
  \And
  Mostafa~Mohsen \\
  Pilotier\\
  Toronto, ON, Canada \\
  \texttt{mostafa@pilotier.com} \\
  \And
  Navya G.~Rao\\
  Pilotier\\
  Toronto, ON, Canada \\
  \texttt{navya@pilotier.com} \\
  \And
  Adam~Forbes\\
  Pilotier\\
  Toronto, ON, Canada \\
  \texttt{adam@pilotier.com} \\
}

\begin{document}

\maketitle

\begin{abstract}
  Autonomous driving has experienced remarkable progress, bolstered by innovations in computational hardware and sophisticated deep learning methodologies. The foundation of these advancements rests on the availability and quality of datasets, which are crucial for the development and refinement of dependable and versatile autonomous driving algorithms. While numerous datasets have been developed to support the evolution of autonomous driving perception technologies, few offer the diversity required to thoroughly test and enhance system robustness under varied weather conditions. Many public datasets lack the comprehensive coverage of challenging weather scenarios and detailed, high-resolution data, which are critical for training and validating advanced autonomous-driving perception models. In this paper, we introduce PLT-D3; a Dynamic-weather Driving Dataset, designed specifically to enhance autonomous driving systems' adaptability to diverse weather conditions. PLT-D3 provides high-fidelity stereo depth and scene flow ground truth data generated using Unreal Engine 5. In particular, this dataset includes synchronized high-resolution stereo image sequences that replicate a wide array of dynamic weather scenarios including rain, snow, fog, and diverse lighting conditions, offering an unprecedented level of realism in simulation-based testing. The primary aim of PLT-D3 is to address the scarcity of comprehensive training and testing resources that can simulate real-world weather variations. Benchmarks have been established for several critical autonomous driving tasks using PLT-D3, such as depth estimation, optical flow and scene-flow to measure and enhance the performance of state-of-the-art models. 
\end{abstract}


\section{Introduction}

Autonomous driving has been a particularly hot topic in the past decade, with numerous contributions at the industry, academia and government levels. While the field of self-driving has seen a great deal of contribution in the form of capital and research, truly autonomous passenger vehicles are still a while away. The challenge mainly stems from the fact that vehicles have the potential to cause lethal damage, hence resulting in the long-tail problem in solving self-driving sufficiently.

The conventional approach to autonomous perception has since been the use of a suite of advanced sensors, including LiDAR and radar, to complement the camera-based vision system \cite{pandharipande2023sensing}. This was primarily due to constraints in compute and lack of robust vision-only perception algorithms. While humans can drive using only their eyesight, such equivalence in the self-driving space has yet to be materialized. Rather, modern camera-only approaches, including Tesla's FSD, Comma's OpenPilot and Wayve's LINGO, all transitioned to an end-to-end approach, bypassing the need for physical scene understanding and reconstruction \cite{cui2024survey, openpilot}. However, to guarantee the reliability of such an end-to-end model, it is imperative that the system, at least internally, has a truly physical understanding of the world.

Nonetheless, the mere problem of utilizing cameras to physically reconstruct the world has been challenging despite showing promising results. In relation to LiDAR, mono and stereo depth camera based networks have shown to have superior density and range while being more robust in adverse weather conditions. Similarly, camera-based optical flow networks provide crisp and high-density motion vectors that dwarfs the resolution of radar sensors. Despite these benefits, numerous challenges remain in utilizing camera images as an alternative to LiDAR and radar, with most SOTA approaches merely using vision to enhance the other sensor measurements \cite{zhang2023perception}. Despite the clear benefits of implementing a vision-based physical perception system, datasets unique to self-driving remain sparse.

The development of datasets tailored for autonomous driving has seen significant advancements over recent years, with several notable contributions aimed at enhancing the robustness and reliability of perception systems. Even though subsequent efforts have expanded the scope and complexity of driving datasets, there is still a lack of diversity of scenes, lighting, and dynamic weather conditions which are crucial for training systems to properly perceive the physical world during adverse conditions.

The fundamentals of physical scene understanding can be represented through depth and scene flow. Forming the basis of the ideal perception sensor (classically modelled using 4D LiDAR or 3D radar), depth combined with scene flow describes the location of every physical object in the scene as well as the direction it is moving. To this extent, we introduction of the PLT-D3 which addresses the aforementioned limitations by offering a comprehensive, high-resolution dataset specifically designed for autonomous driving tasks under diverse lighting and weather conditions. Leveraging the capabilities of Unreal Engine 5 (UE5), PLT-D3 provides synchronized stereo depth and scene flow ground truth data across a wide array of dynamic weather scenarios, including rain, snow, fog, and diverse lighting conditions. This unprecedented level of detail and realism aims to significantly enhance the training and validation of camera-based physical-perception models, thereby pushing the boundaries of camera-based perception technology.

\section{Related Works}
\label{gen_inst}

\subsection{Driving Datasets}
Early datasets, such as KITTI \cite{geiger_kitti}, have been fundamental in establishing benchmarks for tasks like stereo vision, optical flow, and 3D object detection. KITTI's real-world data collection offers high-quality annotations but is limited by its static weather conditions and relatively constrained geographic coverage. Two prominent datasets, BDD100K \cite{yu2020bdd100k} and Cityscapes \cite{cordts2016cityscapes}, provide a diverse set of urban driving scenarios with dense pixel-level annotations, though it primarily focuses on semantic segmentation and does not encompass a wide range of weather conditions. The ApolloScape \cite{huang2018apolloscape} dataset  goes further by incorporating more varied environmental conditions and detailed 3D annotations, yet it still lacks dynamic weather scenarios. The A2D2 \cite{geyer2020a2d2} dataset by Audi includes a broad spectrum of weather and lighting conditions with high-resolution sensor data but concentrates only on semantic segmentation. NuScenes \cite{caesar2020nuscenes} captures diverse urban scenarios, including different weather conditions, but without scene flow and high-resolution depth data.

\subsection{Scene Flow Generation}
In the context of flow, which is a critical component for understanding the motion of objects and the ego-vehicle in a scene. MPI-Sintel \cite{butler2012naturalistic} optical flow dataset, SceneFlow \cite{flyingthings3d} dataset which includes FlyingThings3D, Monkaa, Driving and Spring \cite{mehl2023spring} have been influential but it does not incorporate diverse weather conditions, limiting its applicability for autonomous driving in varied environments. Additionally, the low sample count makes difficult to train perception models the properly generalize to real-world scenarios despite the substantial improvements in 3D graphics simulations.

\subsection{Driving Simulators}
Virtual simulation environments, having been leveraged to address the limitations of real-world data collection, have improved substantially in recent times and enhanced further with the introduction of generative AI techniques. CARLA \cite{dosovitskiy2017carla}, an open-source simulator for autonomous driving research, allows for the generation of diverse driving scenarios under controlled conditions. However, CARLA's weather simulation capabilities, while useful, do not achieve the high-fidelity realism needed for certain advanced perception tasks. The SYNTHIA \cite{ros2016synthia} dataset offers a synthetic collection of images for semantic segmentation and depth prediction, generated under various weather conditions and times of day. Despite its contributions, SYNTHIA's resolution and realism are limited by the capabilities of the simulation engine used.

\subsection{Image Augmentation}
Beyond simulation environments, image augmentation has also been used to effectively diversify and multiply the training dataset. In particular, weather augmentation generation has been used to acquire more diverse conditions and incorporated into the training data, which have been explored in RobustDepth \cite{saunders2023self} and other simulator-based datasets including Virtual-Kitti \cite{gaidon2016virtual}, VIPER \cite{richter2017playing}, Pre-SIL \cite{hurl2019precise}, and GTA-V \cite{richter2016playing}.






\section{The PLT-D3 Dataset}
\label{dataset}

Focusing on a camera-only approach to physical perception of autonomous driving, we present the PLT-D3 dataset based on a front-facing stereo camera setup. Rendered using UE5 to provide exceptional photo-realism and diverse driving scenarios, the dataset contains high-quality high-resolution images along with ground truth depth, disparity, optical flow and scene flow. \footnote{Link to the dataset: \href{https://doi.org/10.7910/DVN/36SQKM}{https://doi.org/10.7910/DVN/36SQKM}.}

\subsection{Driving Scenarios}

\begin{figure}
  \label{fig:dynamic_samples}
  \centering
  \includegraphics[scale=0.1]{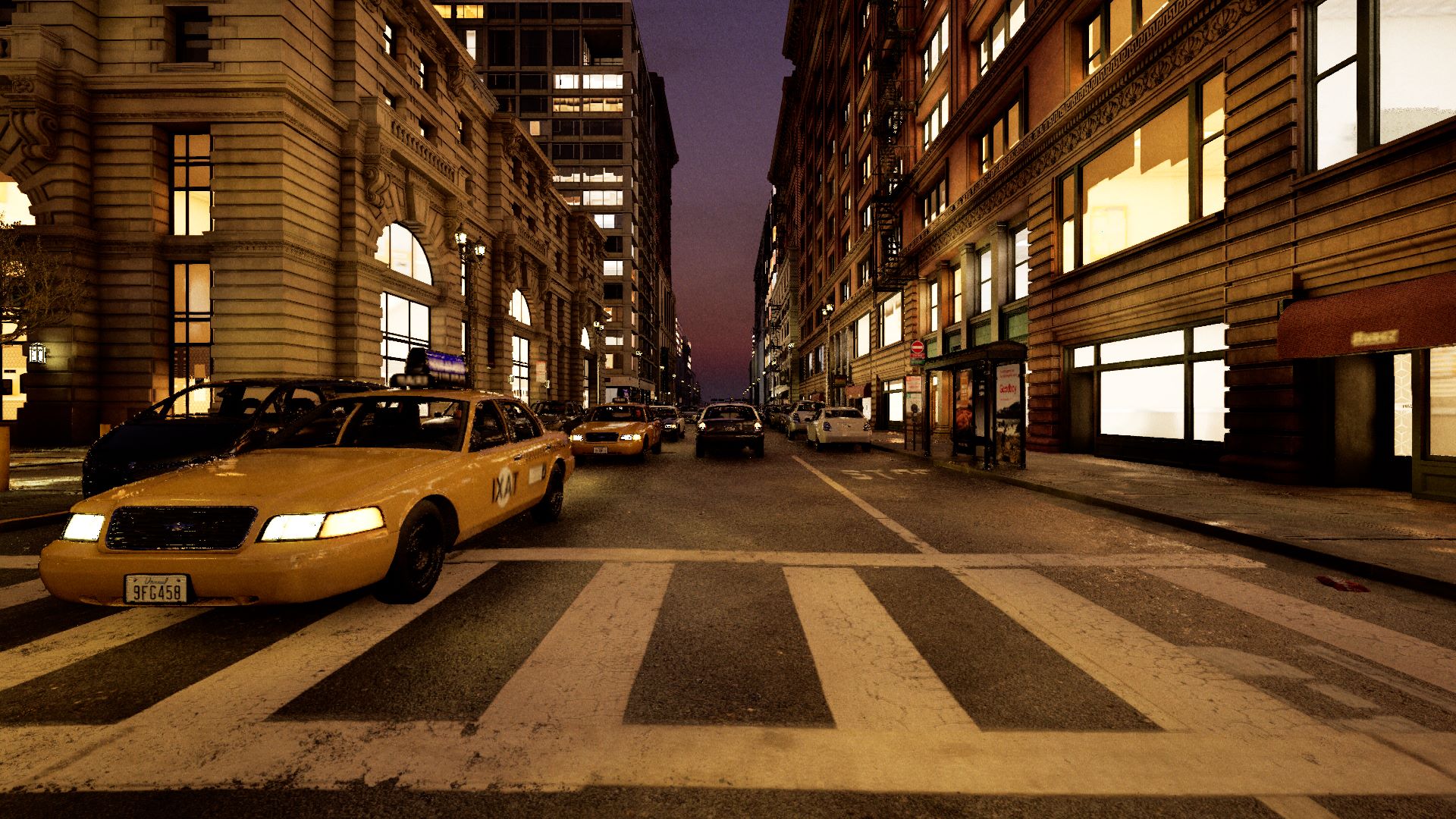}
  \includegraphics[scale=0.1]{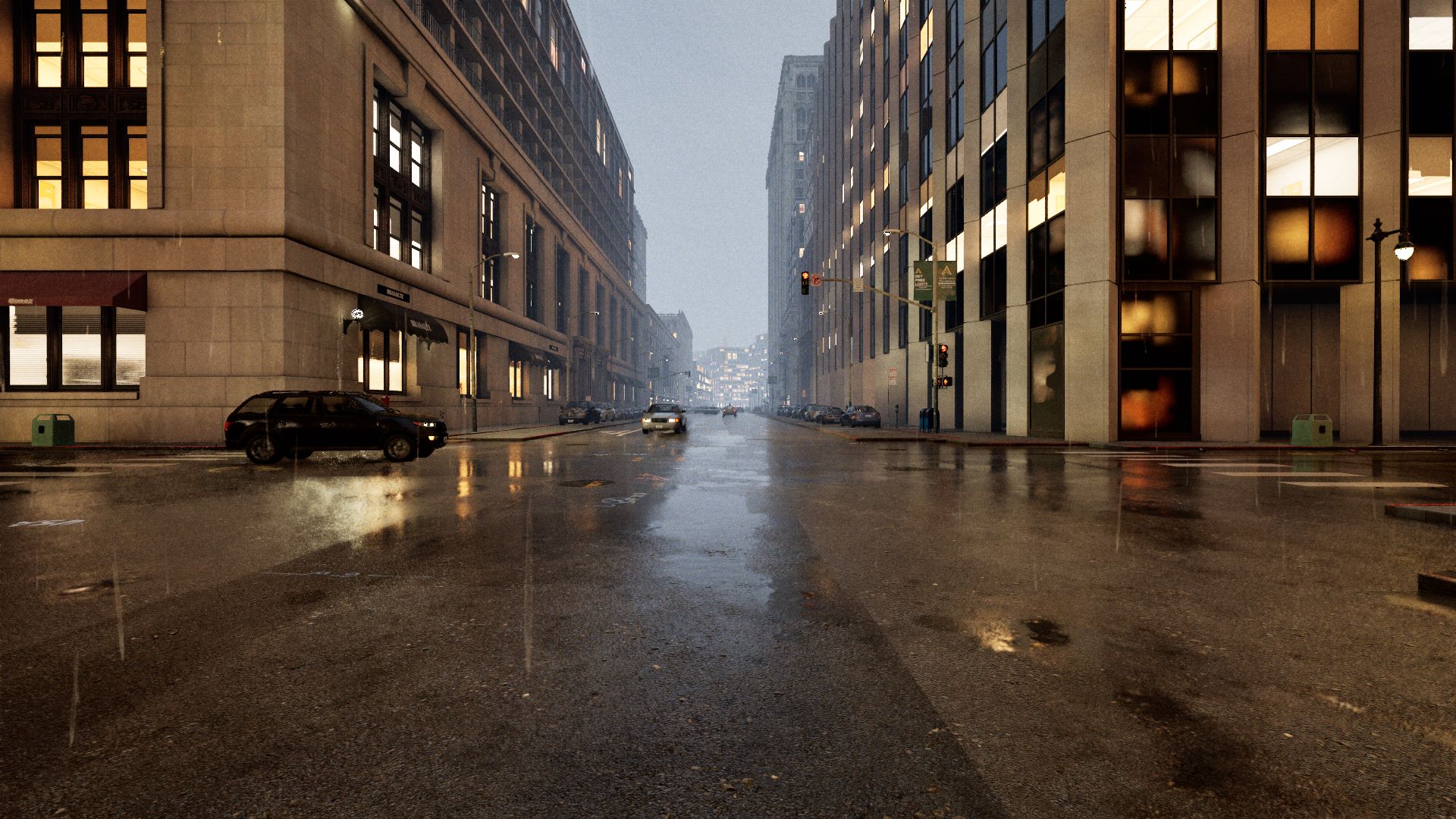}
  \includegraphics[scale=0.1]{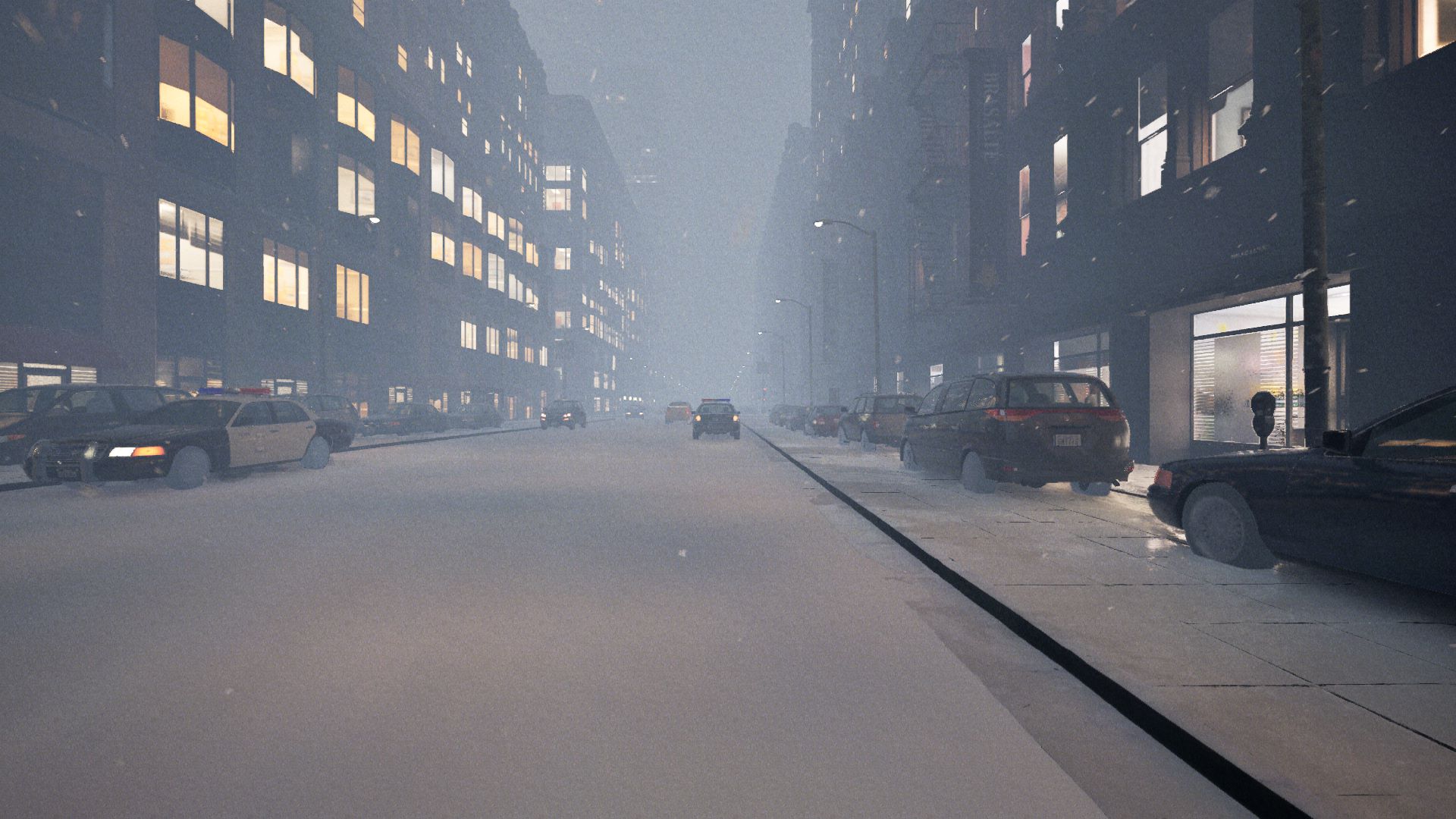}
  \includegraphics[scale=0.1]{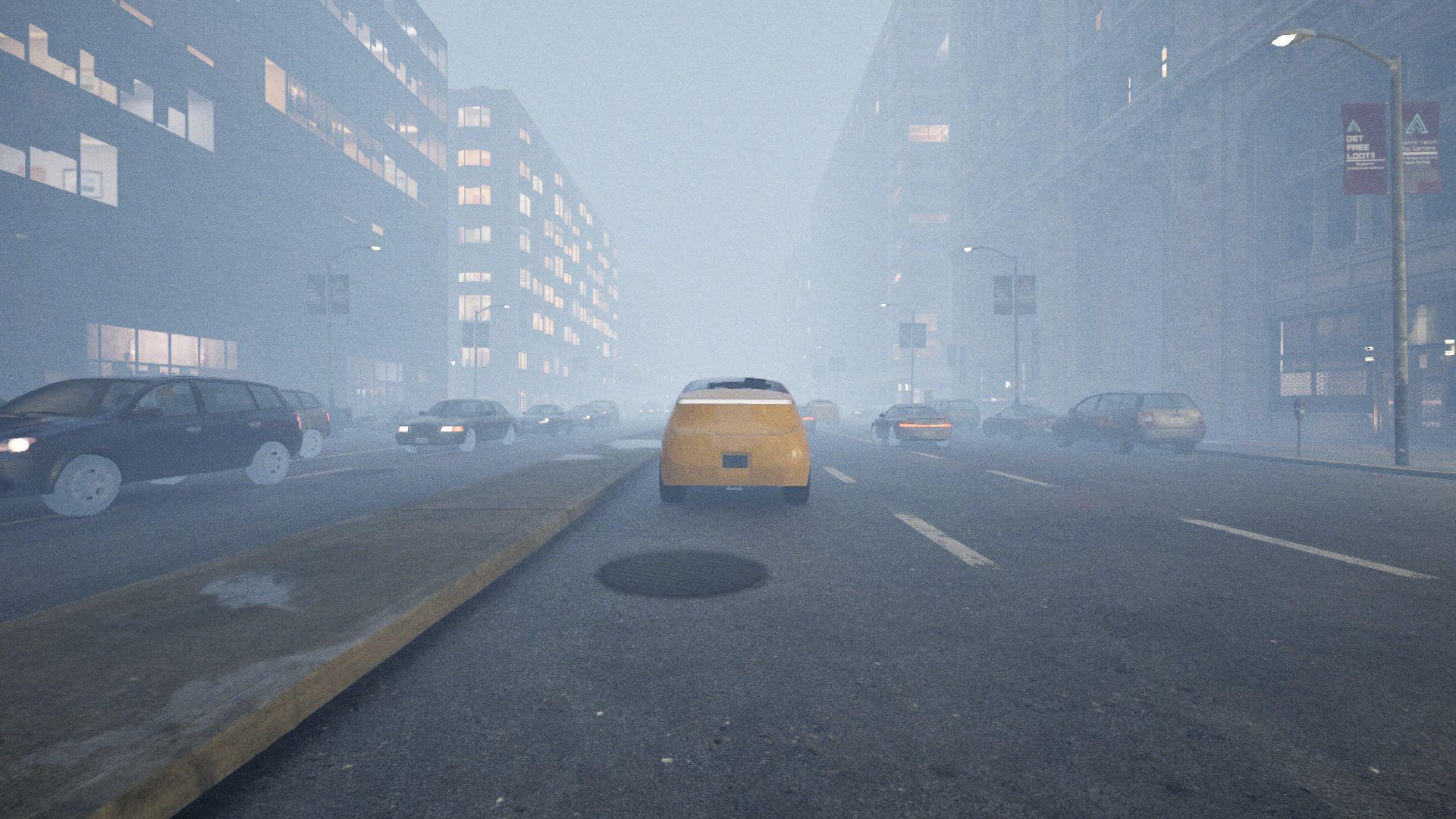}
  \includegraphics[scale=0.1]{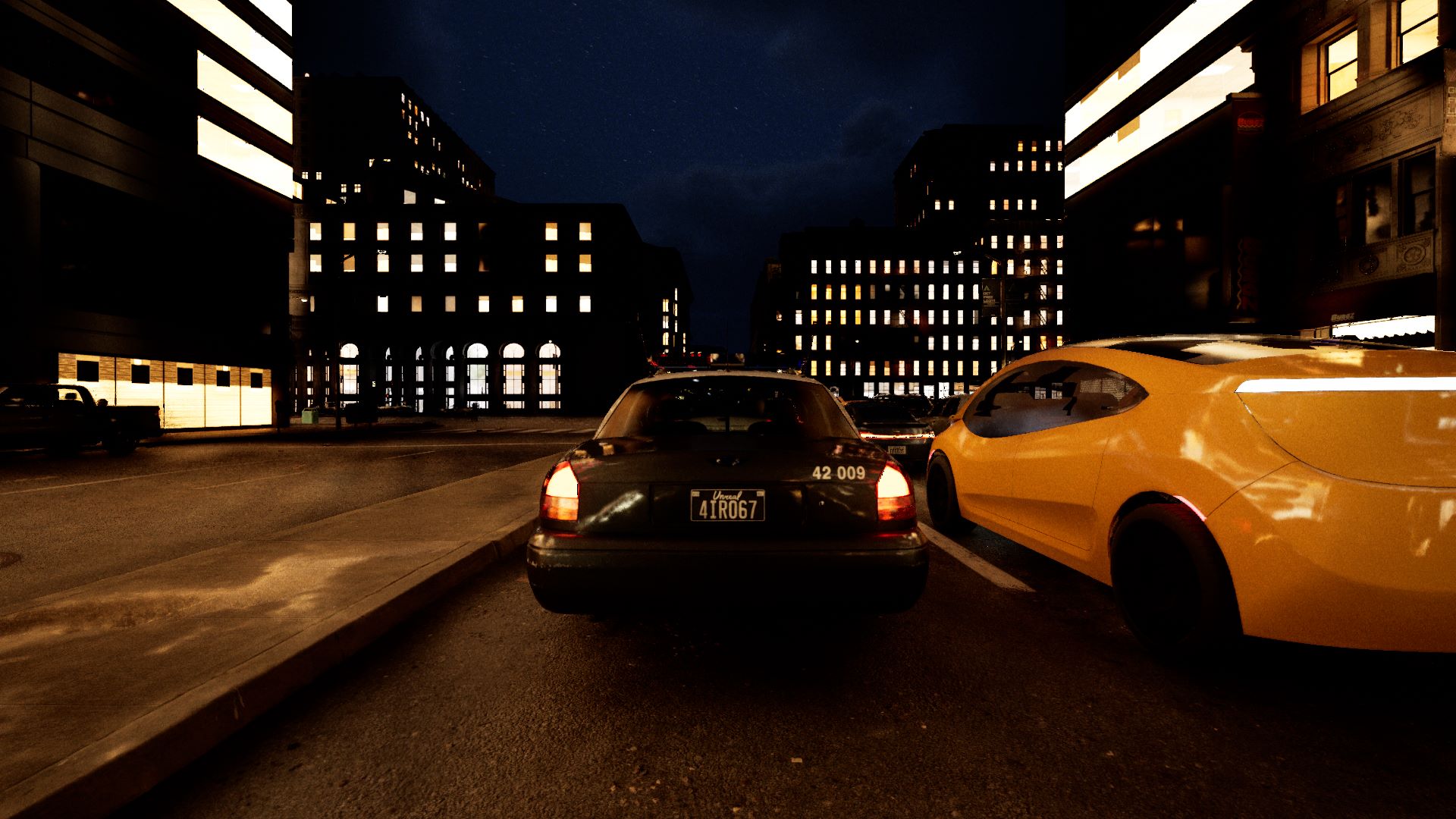}
  \includegraphics[scale=0.1]{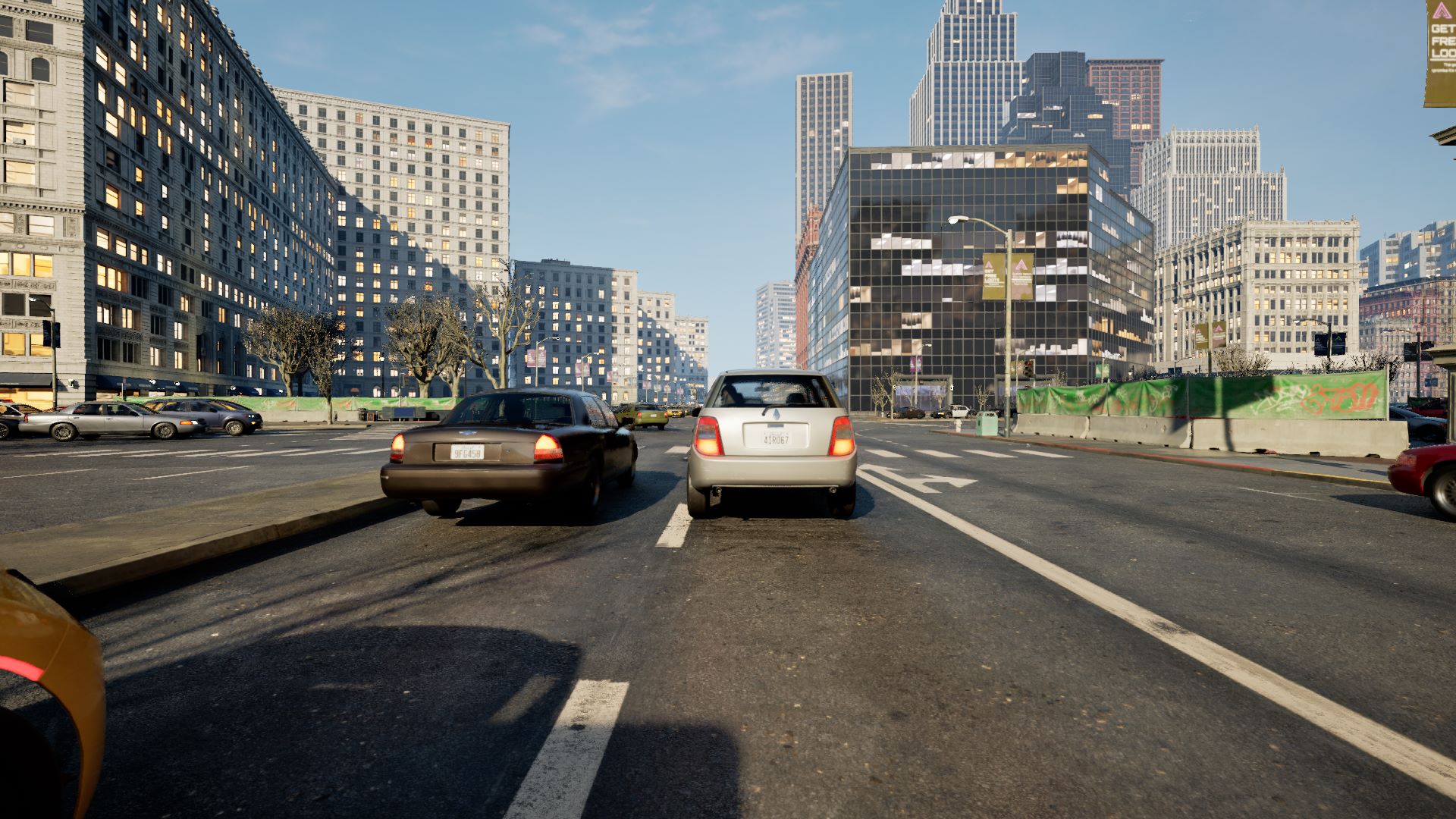}
  \includegraphics[scale=0.1]{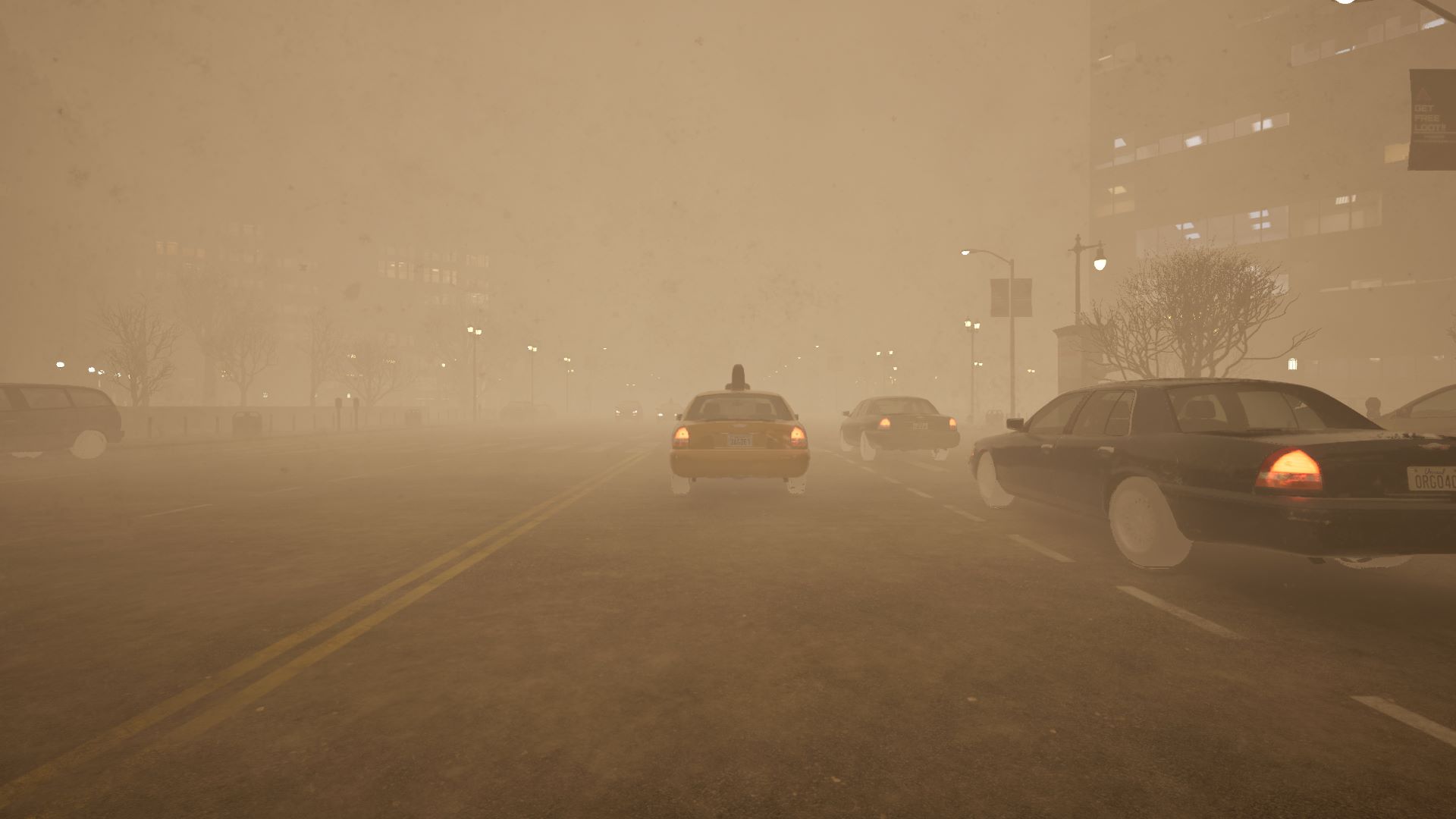}
  \includegraphics[scale=0.1]{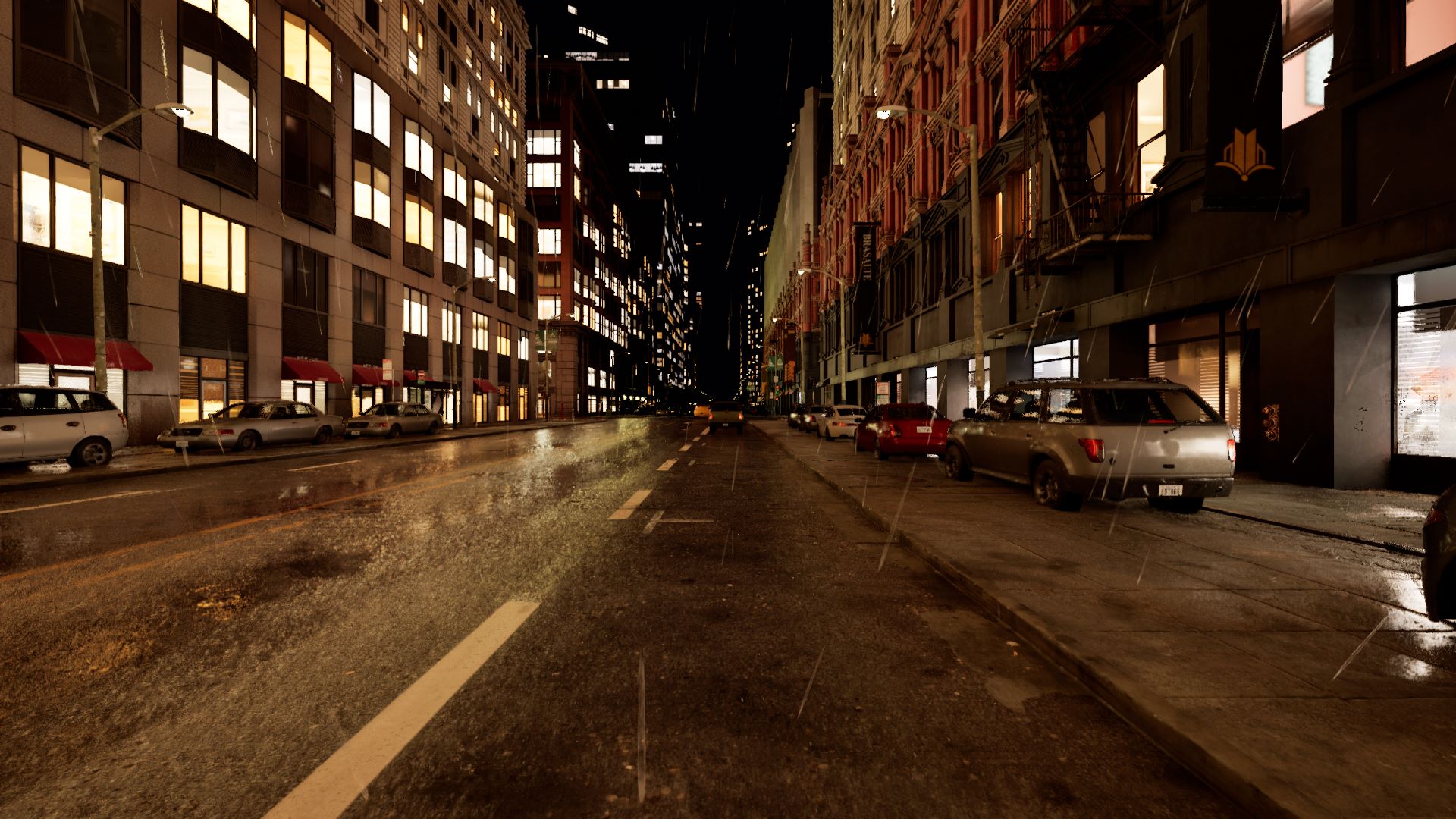}
  \includegraphics[scale=0.1]{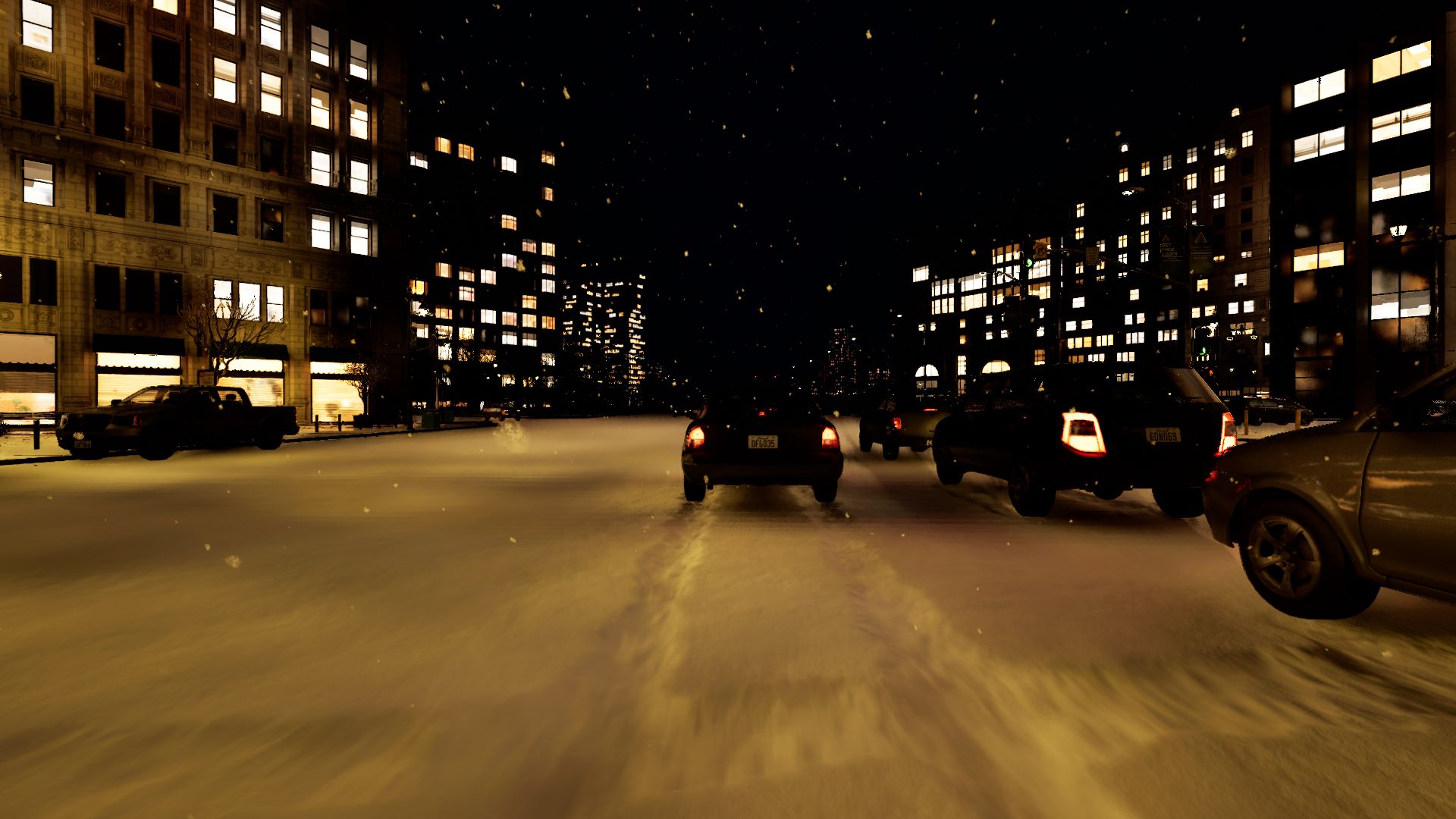}
  \includegraphics[scale=0.1]{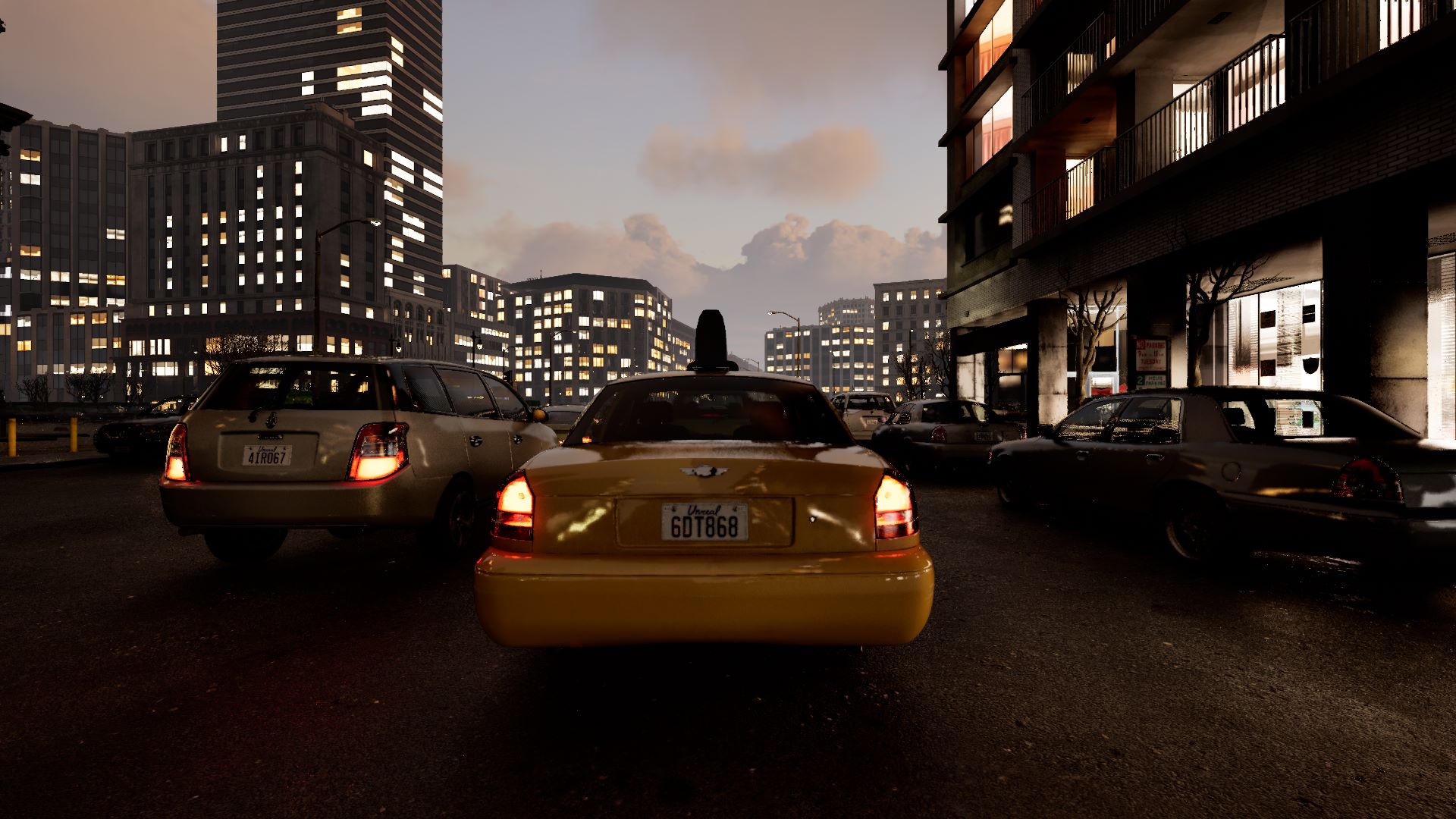}
  \caption{Sample RGB images from PLT-D3. From top-to-bottom (left-to-right): (sun \& rain), (snow \& fog), (cloudy \& sunny), (dust storm \& rain) and (snow \& partly cloudy).}
\end{figure}

When discussing the edge-cases in camera-based autonomous driving perception, two particular factors appear to be mentioned often: lighting condition and weather. Despite occurring daily, excessive low-light and glare often have a destructive effect on physical perception algorithms. Likewise, various weather conditions from rain to snow and overcast to foggy all impact the robustness of camera-base perception. Hence, the PLT-D3 dataset contains a proportionate mix of diverse lighting and weather conditions as shown in Fig. \ref{fig:dynamic_samples}.

\subsubsection{Dynamic Lighting}
Lighting plays a major role in this dataset in three main ways. Firstly, the time of day is altered, consisting of sunrise, midday, twilight and midnight. Secondly, a diverse set of buildings are scattered throughout the simulation world, from high-rise to short-story. Within these diverse set of buildings, some emanate bright lights through their glass window while others have limited illumination. In addition, lights from various types of lamp posts and road structure impact the final camera lighting. Lastly, although an effect of weather, lightning occurring from stormy weathers (snow, rain, etc.) produce bright flashes primarily in the night, contributing to the diverse lighting conditions.

\subsubsection{Dynamic Weather}
Similar to lighting, there are five main weather types proportionately present in the dataset. These include clear sky, rain, snow, fog and smog. Within these major categories are subsets as well, which are explained in Table \ref{tab:weather}. In this dataset, while a random assortment of all the subcategories have been utilized, only the major category is identified within a subset of the dataset. Performing only high-level categorization appears to be sufficient to train perception models for these adverse models, as sub-categorization does not appear to aid in perception generalization.

\begin{table}
  \caption{Dynamic weather in PLT-D3 Simulator.}
  \label{tab:weather}
  \centering
  \begin{tabular}{lccccc}
    \toprule
    
    Categories  & Mild & Average & Extreme \\
    \midrule
    Cloud & Clear Sky   & Partly Cloudy & Overcast \\
    Rain & Drizzle              & Rain              & Thunderstorm \\
    Snow & Flurries              & Snow              & Blizzard \\
    Fog & -               & Fog               & - \\
    Smog  & -               & Smog               & Dust storm \\
    
    
    \bottomrule
  \end{tabular}
\end{table}


\subsection{Simulation Engine}
\label{sec:simulation_engine}

UE5 was used as the simulation engine for the generation of the dataset.  The simulated level is a section of Brooklyn, centered around Grand Army Plaza and spanning approximately six by eight kilometers. 3D Model tile sets, vehicle models, and pedestrian models from the "City Sample" product from the Epic Games Marketplace were used \cite{EpicCitySample}. \footnote{These assets are free to be used both commercially and non-commercially, so long as they are used with Unreal Engine, since these assets are elements created by Epic Games \cite{epic2024lisence}.}

There are two primary issues faced when tasked with generating stereo image sets from UE5 with both depth and optical flow ground truths.  The first issue is the lack of support for multi-camera rendering within vanilla UE5; the second issue is the lack of support for correct optical flow outputs. \footnote{The erroneous code can be seen in their \href{https://forums.unrealengine.com/t/problem-interpreting-scenetexture-velocity-data/463593/6}{forum post}; the issue still remains.}  While the use of multiple cameras and the output of optical flow ground truths is well supported in traditional animation software like Blender, using a traditional animation software loses out several of the advantages of using UE5, specifically the speed at which high fidelity datasets can be generated, and the much greater support for procedural generation and independent actors. With the goal to focus on a procedural generated city with intelligent traffic for ultra-realism, UE5 was seen as the best option despite the aforementioned shortcomings. The simulator hence made use of a custom engine patch that enabled synchronized multi-camera setup and corrected optical flow data.

\subsection{Dataset Structure}
\begin{figure}
  \label{fig:samples}
  \centering
  \includegraphics[scale=0.068]{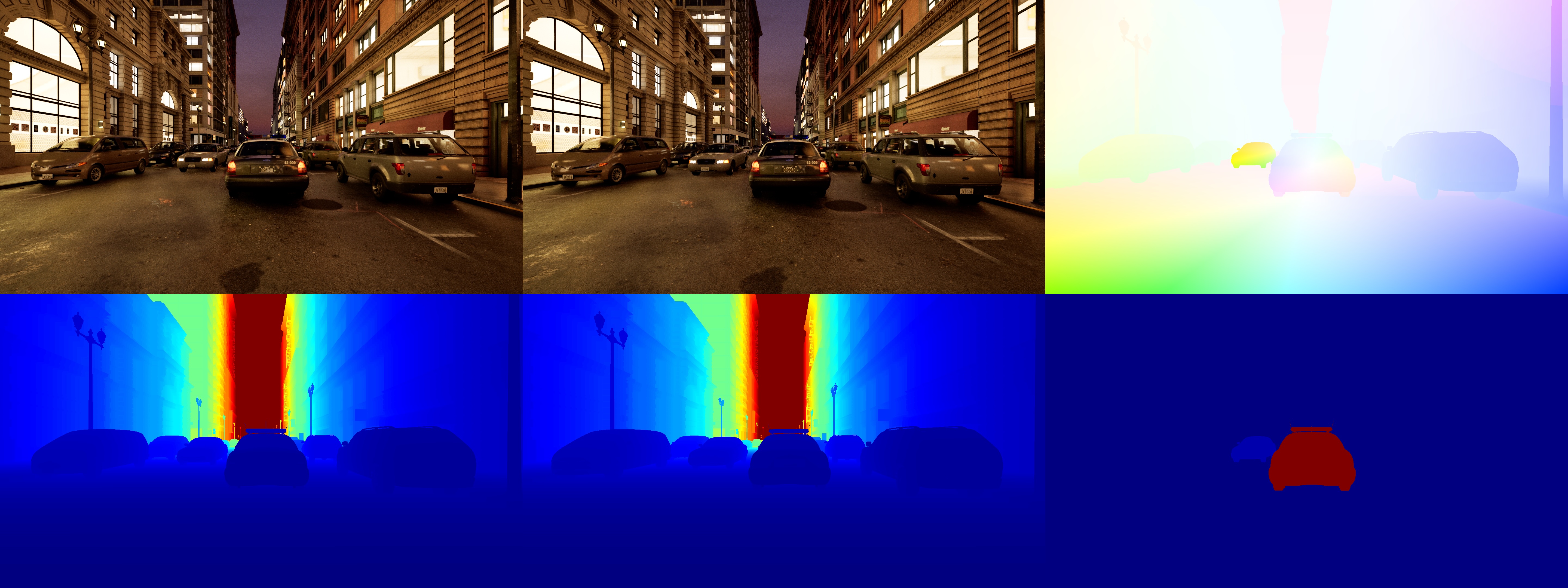}
  \caption{Ground truth visualization from one sample. From left-to-right then top-to-bottom: left RGB, right RGB, optical flow, left depth, right depth, and delta disparity.}
\end{figure}

The dataset contains 10 unique driving scenarios comprised of a particular time of day and weather condition. Each scenario has a number of scenes ranging from 5 to 20, where each scene contains exactly 21 samples spaced at 10Hz (100ms apart). Within each sample are Full-HD (1920x1080p) left and right images (12cm baseline) along with ground truth optical flow, depth, delta disparity and camera metadata all corresponding to the left camera. Visualization of a particular sample is shown in Fig. \ref{fig:samples}.

\subsection{Limitations}
\label{sec:limitations}

While the PLT-D3 dataset marks a significant advancement in simulating dynamic driving environments for autonomous driving research, it does come with a few limitations. The dataset created using UE5 cannot fully replicate the complexities and unpredictable nature of real-world conditions, like precise material properties and human behaviors. Additionally, the dataset focuses on stereo camera data and does not include other sensor data like LiDAR and radar, which are essential for various sensor fusion algorithms in autonomous vehicles.. The high-resolution data and detailed annotations require substantial computational resources, potentially limiting use to those with access to high-performance computing systems. These factors should be considered when using the PLT-D3 dataset to ensure its applications are appropriate and its limitations are understood.


\section{Benchmarks}
\label{sec:benchmarks}

In the structuring of our dataset, we divided it into 10 distinct scenes, resulting in 3,465 stereo frames allocated to the training set and 361 stereo frames designated for the test set. We provide complete access to the training split data; however, for the test split, only the images are made publicly available while the ground truth files are withheld, aligning with standard practices \cite{baker_database} \cite{butler2012naturalistic} \cite{dalvand2016fast} \cite{geiger_kitti} \cite{Menze_2015_CVPR} \cite{Ranjan_2020} \cite{richter2017playing} \cite{scharstein} in the field. This method of data distribution helps maintain the confidentiality of the test data and enhances the robustness of the evaluation process by preventing overfitting and ensuring that the methods are genuinely capable of generalizing to unseen data. Additionally, this structured approach facilitates a fair comparison of different methodologies by ensuring that all participants are evaluated under consistent conditions.

\subsection{Evaluation Metrics}
\label{sec:metrics}

In the evaluation of scene flow, a variety of metrics are employed to assess error measurement, crucial for applications in optical flow, stereo and scene-flow methods. Scene flow errors are quantified based on deviations from the ground truth, specifically using thresholds of 1 pixel (px) and 3 pixels (px), indicating the maximum allowable displacement error. This is essential for autonomous systems given the variability and complexity of these environments. For optical flow analysis, traditional metrics such as 1px, 3px, and 5px error rates along with End-Point Error (EPE) and Root Mean Square Error (RMSE) are used to benchmark as these metrics are proven to effectively measure the accuracy of models. EPE quantifies the average Euclidean distance between the predicted and actual end points of the motion vector. RMSE measures the square root of the average squared differences between predicted and actual values, providing a standard for assessing the overall error magnitude.
In the assessment of disparity, which refers to the difference in coordinates of similar features within two stereo images, we use metrics such as Bad 1.0, Bad 2.0. 'Bad 1.0' and 'Bad 2.0' refer to the percentage of pixels with a disparity error greater than 1px and 2px respectively, offering a discrete measure of disparity accuracy. The aforementioned metrics and their implications are systematically presented in a tabular format, detailing performance across different parts of the scene. This structured presentation not only aids in comprehending the models effectiveness but also highlights areas requiring improvement. The evaluation is conducted on a computing system utilizing an x86\_64 architecture, equipped with an AMD Ryzen 9 7950X 16-Core Processor and an NVIDIA RTX 4090 graphics card.

\subsection{Results}
\label{sec:results}

In our evaluation, we compared 13 state-of-the-art methods across three different categories: stereo depth (3 methods), optical flow (7 methods), and scene flow (3 methods). The results are detailed in Tables \ref{disp-table}, \ref{of-table}, and \ref{sf-table}, respectively. Each of these models was selected based on the availability of their code and pre-trained checkpoints. We utilized models that had been fine-tuned on the KITTI dataset, a standard benchmark for driving datasets, to ensure relevance to our domain. The use of pre-trained models offers preliminary insights into the generalization capabilities of these methods across different settings. However, it is important to note that these results should be viewed as initial indicators rather than definitive assessments of each method's efficacy. To better understand the full potential and effectiveness of these approaches, we encourage researchers to submit versions of their methods that have been fine-tuned specifically for our benchmark. This will allow for a more accurate and comprehensive evaluation of their performance in context-specific scenarios.
\subsubsection{Stereo Depth}

For stereo depth estimation, the performance of the evaluated models significantly surpasses that observed in both optical flow and scene flow evaluations. CREStereo \cite{crestereo}, which has been trained on a large and diverse dataset, demonstrates notable accuracy on unseen data, showing its robustness and generalizability. Other methods, such as HITNET \cite{hitnet}, while not achieving the same level of performance initially, exhibit potential for substantial improvement. By training or fine-tuning these models on the PLT-D3 dataset, it is anticipated that their accuracy and adaptability to diverse real-world conditions will be markedly enhanced. This suggests a promising avenue for advancing stereo estimation capabilities through dataset-specific optimization.
\subsubsection{Optical Flow}

For optical flow estimation, our analysis reveals that while state-of-the-art (SOTA) models generally exhibit robust performance, they also display significant error margins, particularly in regions requiring 1-pixel accuracy. Notably, areas such as the sky demonstrate substantial inaccuracies. Among the evaluated models, RAFT \cite{teed2020raft}, which serves as a common backbone for many recent optical flow methods, exhibits the highest error margins. Conversely, FlowNet2 \cite{flownet2} and IRR \cite{irr} outperform other models, suggesting their architectures are better generalized in handling high-resolution images within this dataset. This observation underscores the importance of architectural considerations when tackling complex datasets that demand high levels of precision.

\subsubsection{Scene Flow}
For scene flow estimation, our evaluation involves three methods originally trained on the FlyingThings3D \cite{flyingthings3d} dataset and fine-tuned on KITTI \cite{geiger_kitti}. The results reveal a tendency for these methods to overfit to the KITTI \cite{geiger_kitti} dataset, resulting in diminished performance when tested on unseen data. This pattern underscores a critical limitation in current training approaches and strongly highlights the necessity of the PLT-D3 dataset. The PLT-D3 dataset is vital resource for pushing the boundaries of scene flow research, providing a crucial test-bed for developing methodologies that are not only accurate but also highly generalizable to diverse real-world scenarios.

\begin{table}
  \caption{Stereo depth evaluation results on PLT-D3. We show the Bad Pixel Rate at thresholds of 1.0 and 2.0 (Bad 1.0, Bad 2.0), the Average Endpoint Error (EPE), and the Root Mean Square Error (RMSE), with sub-rankings high/low. }
  \label{disp-table}
  \centering
  \begin{tabular}{lccccc}
    \toprule
    Model         & Bad 1.0 (\%) & Bad 2.0 (\%) & EPE   & RMSE \\
    \midrule
    LEAStereo \cite{leastereo}     & 70.66               & 41.21               & 7.47  & 12.52 \\
    HITNET (M) \cite{hitnet}    & 25.06              & 19.01              & 2.81  & 6.04 \\
    HITNET (E)  \cite{hitnet}  & 21.35              & 15.27              & 2.92  & 5.67 \\
    CREStereo \cite{crestereo}     & 7.86               & 4.11               & 0.68  & 1.93 \\
    
    \bottomrule
  \end{tabular}
\end{table}

\begin{table}
  \caption{Optical Flow evaluation results on PLT-D3. We demonstrate the Average Endpoint Error (EPE) for both 2D and show the performance for the pixel accuracy metrics \( \delta_{2D} <\) 1px , 3px, 5px outlier rate with sub-rankings high/low. }
  \label{of-table}
  \centering
  \begin{tabular}{lccccc}
    \toprule
    Method       & Resolution  & EPE    & \( \delta_{2D} <\) 1px    & \( \delta_{2D} <\) 3px    & \( \delta_{2D} <\) 5px    \\
    \midrule
    RAFT \cite{teed2020raft}        & 1080 x 1920 & 51.22  & 0.45   & 0.56   & 0.62   \\
    GMA   \cite{gma}       & 1080 x 1920 & 29.97  & 0.47   & 0.58   & 0.64   \\
    PWCNet   \cite{pwcnet}    & 1080 x 1920 & 29.16  & 0.44   & 0.58   & 0.65   \\
    FlowFormer \cite{floformer}  & 540 x 960   & 25.89  & 0.38   & 0.48   & 0.55   \\
    LiteFlowNet2 \cite{liteflownet2} & 1080 x 1920 & 21.87  & 0.47   & 0.60   & 0.66   \\
    FlowNet2   \cite{flownet2}  & 1080 x 1920 & 16.97  & 0.48   & 0.61   & 0.69   \\
    IRR \cite{irr}       & 1080 x 1920 & 16.85  & 0.46   & 0.59   & 0.66   \\
    \bottomrule
  \end{tabular}
\end{table}

\begin{table}
  \caption{Scene Flow Evaluation Results on PLT-D3. We demonstrate the average endpoint error (EPE) for both 2D and 3D perspectives. Additionally, we show the performance for the pixel accuracy metrics \( \delta_{2D} <\) 1px and \( \delta_{2D} <\) 3px, along with the 3D distance accuracy metrics \( \delta_{3D} < .05\) (\%) and \( \delta_{3D} < .10\) (\%) with sub-rankings high/low.}
  
  \label{sf-table}
  \centering
   \begin{tabular}{lcccccc}
    \toprule
    Model       & \multicolumn{3}{c}{2D Metrics} & \multicolumn{3}{c}{3D Metrics} \\
    \cmidrule(lr){2-4} \cmidrule(lr){5-7}
                 & EPE 2D & \( \delta_{2D} <\) 1px & \( \delta_{2D} <\) 3px & EPE 3D & \( \delta_{3D} < .05\) (\%) & \( \delta_{3D} < .10\) (\%) \\
    \midrule
    RAFT-3D \cite{raft3d}     & 14.43  & 0.3815 & 0.4850 & 32.06 & 32.99 & 34.03 \\
    ScaleRAFT \cite{scaleraft}   & 9.84   & 0.3777 & 0.4828 & 32.02 & 27.60 & 30.71 \\
    MFuse \cite{mfuse}      & 10.51  & 0.3810 & 0.4861 & 10.51 & 29.18 & 30.89 \\
    \bottomrule
  \end{tabular}
\end{table}


\section{Conclusion}
\label{sec:conclusion}
In this paper, we introduced the PLT-D3 dataset, a dynamic, high-fidelity simulation dataset designed specifically to advance the capabilities of autonomous driving systems under diverse weather and lighting conditions. By leveraging the rendering capabilities of UE5, PLT-D3 provides an unparalleled level of realism in stereo depth, optical flow and scene flow data, essential for training and validating advanced perception models. Our dataset addresses a critical gap in existing autonomous driving datasets, which often lack comprehensive coverage of adverse weather scenarios and dynamic lighting conditions. Through systematic benchmarks established using PLT-D3, we have demonstrated the potential of this dataset to significantly enhance the performance of state-of-the-art autonomous driving algorithms. The results indicate that while current models perform reasonably well on standard datasets like KITTI, they struggle to generalize to the complex scenarios presented in PLT-D3, underscoring the necessity of datasets that include challenging environmental conditions. Initial results of the 13 non-finetuned baselines showed that the PLT-D3 dataset not only serves as a vital tool for developing more accurate and reliable perception systems but also as a benchmark for future research in autonomous driving. As autonomous driving technology continues to proliferate, datasets like PLT-D3 will be crucial in pushing the boundaries of what is possible, moving us closer to the realization of fully autonomous systems capable of operating safely and efficiently in any environmental condition.









\begin{ack}
This research was partially funded by the National Research Council of Canada Industrial Research Assistance Program (NRC IRAP).
\end{ack}


\printbibliography

@article{dalvand2016fast,
  title={Fast vision-based catheter 3D reconstruction},
  author={Dalvand, Mohsen Moradi and Nahavandi, Saeid and Howe, Robert D},
  journal={Physics in Medicine \& Biology},
  volume={61},
  number={14},
  pages={5128},
  year={2016},
  publisher={IOP Publishing}
}

@INPROCEEDINGS{baker_database,
  author={Baker, Simon and Roth, Stefan and Scharstein, Daniel and Black, Michael J. and Lewis, J.P. and Szeliski, Richard},
  booktitle={2007 IEEE 11th International Conference on Computer Vision}, 
  title={A Database and Evaluation Methodology for Optical Flow}, 
  year={2007},
  volume={},
  number={},
  pages={1-8},
  doi={10.1109/ICCV.2007.4408903}}

@INPROCEEDINGS{geiger_kitti,
  author={Geiger, Andreas and Lenz, Philip and Urtasun, Raquel},
  booktitle={2012 IEEE Conference on Computer Vision and Pattern Recognition}, 
  title={Are we ready for autonomous driving? The KITTI vision benchmark suite}, 
  year={2012},
  volume={},
  number={},
  pages={3354-3361},
  doi={10.1109/CVPR.2012.6248074}}

@InProceedings{Menze_2015_CVPR,
author = {Menze, Moritz and Geiger, Andreas},
title = {Object Scene Flow for Autonomous Vehicles},
booktitle = {Proceedings of the IEEE Conference on Computer Vision and Pattern Recognition (CVPR)},
month = {6},
year = {2015}
}

@article{Ranjan_2020,
   title={Learning Multi-human Optical Flow},
   volume={128},
   ISSN={1573-1405},
   url={http://dx.doi.org/10.1007/s11263-019-01279-w},
   DOI={10.1007/s11263-019-01279-w},
   number={4},
   journal={International Journal of Computer Vision},
   publisher={Springer Science and Business Media LLC},
   author={Ranjan, Anurag and Hoffmann, David T. and Tzionas, Dimitrios and Tang, Siyu and Romero, Javier and Black, Michael J.},
   year={2020},
   month=jan, pages={873–890} }

@INPROCEEDINGS{scharstein,
  author={Scharstein, D. and Szeliski, R. and Zabih, R.},
  booktitle={Proceedings IEEE Workshop on Stereo and Multi-Baseline Vision (SMBV 2001)}, 
  title={A taxonomy and evaluation of dense two-frame stereo correspondence algorithms}, 
  year={2001},
  volume={},
  number={},
  pages={131-140},
  doi={10.1109/SMBV.2001.988771}}

@misc{epic2024lisence, title={Unreal Engine Marketplace Support}, url={https://marketplacehelp.epicgames.com/s/?language=en_US}, journal={Marketplace support}, author={Epic Games}}

@misc{EpicCitySample, title={City sample - UE marketplace}, url={https://www.unrealengine.com/marketplace/en-US/product/city-sample}, journal={Unreal Engine Marketplace}, author={Epic Games}}

@misc{hitnet,
      title={HITNet: Hierarchical Iterative Tile Refinement Network for Real-time Stereo Matching}, 
      author={Vladimir Tankovich and Christian Häne and Yinda Zhang and Adarsh Kowdle and Sean Fanello and Sofien Bouaziz},
      year={2023},
      eprint={2007.12140},
      archivePrefix={arXiv},
      primaryClass={cs.CV}
}

@misc{crestereo,
      title={Practical Stereo Matching via Cascaded Recurrent Network with Adaptive Correlation}, 
      author={Jiankun Li and Peisen Wang and Pengfei Xiong and Tao Cai and Ziwei Yan and Lei Yang and Jiangyu Liu and Haoqiang Fan and Shuaicheng Liu},
      year={2022},
      eprint={2203.11483},
      archivePrefix={arXiv},
      primaryClass={cs.CV}
}

@misc{gma,
      title={Learning to Estimate Hidden Motions with Global Motion Aggregation}, 
      author={Shihao Jiang and Dylan Campbell and Yao Lu and Hongdong Li and Richard Hartley},
      year={2021},
      eprint={2104.02409},
      archivePrefix={arXiv},
      primaryClass={cs.CV}
}

@misc{pwcnet,
      title={PWC-Net: CNNs for Optical Flow Using Pyramid, Warping, and Cost Volume}, 
      author={Deqing Sun and Xiaodong Yang and Ming-Yu Liu and Jan Kautz},
      year={2018},
      eprint={1709.02371},
      archivePrefix={arXiv},
      primaryClass={cs.CV}
}

@misc{floformer,
      title={FlowFormer: A Transformer Architecture for Optical Flow}, 
      author={Zhaoyang Huang and Xiaoyu Shi and Chao Zhang and Qiang Wang and Ka Chun Cheung and Hongwei Qin and Jifeng Dai and Hongsheng Li},
      year={2022},
      eprint={2203.16194},
      archivePrefix={arXiv},
      primaryClass={cs.CV}
}

@misc{liteflownet2,
      title={A Lightweight Optical Flow CNN - Revisiting Data Fidelity and Regularization}, 
      author={Tak-Wai Hui and Xiaoou Tang and Chen Change Loy},
      year={2020},
      eprint={1903.07414},
      archivePrefix={arXiv},
      primaryClass={cs.CV}
}

@misc{flownet2,
      title={FlowNet 2.0: Evolution of Optical Flow Estimation with Deep Networks}, 
      author={Eddy Ilg and Nikolaus Mayer and Tonmoy Saikia and Margret Keuper and Alexey Dosovitskiy and Thomas Brox},
      year={2016},
      eprint={1612.01925},
      archivePrefix={arXiv},
      primaryClass={cs.CV}
}

@misc{raft3d,
      title={RAFT-3D: Scene Flow using Rigid-Motion Embeddings}, 
      author={Zachary Teed and Jia Deng},
      year={2021},
      eprint={2012.00726},
      archivePrefix={arXiv},
      primaryClass={cs.CV}
}

@misc{scaleraft,
      title={High Resolution Multi-Scale RAFT (Robust Vision Challenge 2022)}, 
      author={Azin Jahedi and Maximilian Luz and Lukas Mehl and Marc Rivinius and Andrés Bruhn},
      year={2022},
      eprint={2210.16900},
      archivePrefix={arXiv},
      primaryClass={cs.CV}
}

@inproceedings{mfuse,
  title={{M-FUSE}: Multi-frame Fusion for Scene Flow Estimation},
  author={Mehl, Lukas and Jahedi, Azin and Schmalfuss, Jenny and Bruhn, Andr{\'e}s},
  booktitle={Proc. Winter Conference on Applications of Computer Vision (WACV)},
  year={2023}
}

@article{irr,
  title={Iterative Residual Refinement for Joint Optical Flow and Occlusion Estimation},
  author={Junhwa Hur and Stefan Roth},
  journal={2019 IEEE/CVF Conference on Computer Vision and Pattern Recognition (CVPR)},
  year={2019},
  pages={5747-5756},
  url={https://api.semanticscholar.org/CorpusID:131773871}
}

@inproceedings{caesar2020nuscenes,
  title={nuscenes: A multimodal dataset for autonomous driving},
  author={Caesar, Holger and Bankiti, Varun and Lang, Alex H and Vora, Sourabh and Liong, Venice Erin and Xu, Qiang and Krishnan, Anush and Pan, Yu and Baldan, Giancarlo and Beijbom, Oscar},
  booktitle={Proceedings of the IEEE/CVF conference on computer vision and pattern recognition},
  pages={11621--11631},
  year={2020}
}

@inproceedings{cordts2016cityscapes,
  title={The cityscapes dataset for semantic urban scene understanding},
  author={Cordts, Marius and Omran, Mohamed and Ramos, Sebastian and Rehfeld, Timo and Enzweiler, Markus and Benenson, Rodrigo and Franke, Uwe and Roth, Stefan and Schiele, Bernt},
  booktitle={Proceedings of the IEEE conference on computer vision and pattern recognition},
  pages={3213--3223},
  year={2016}
}

@inproceedings{huang2018apolloscape,
  title={The apolloscape dataset for autonomous driving},
  author={Huang, Xinyu and Cheng, Xinjing and Geng, Qichuan and Cao, Binbin and Zhou, Dingfu and Wang, Peng and Lin, Yuanqing and Yang, Ruigang},
  booktitle={Proceedings of the IEEE conference on computer vision and pattern recognition workshops},
  pages={954--960},
  year={2018}
}

@article{geyer2020a2d2,
  title={A2d2: Audi autonomous driving dataset},
  author={Geyer, Jakob and Kassahun, Yohannes and Mahmudi, Mentar and Ricou, Xavier and Durgesh, Rupesh and Chung, Andrew S and Hauswald, Lorenz and Pham, Viet Hoang and M{\"u}hlegg, Maximilian and Dorn, Sebastian and others},
  journal={arXiv preprint arXiv:2004.06320},
  year={2020}
}

@inproceedings{butler2012naturalistic,
  title={A naturalistic open source movie for optical flow evaluation},
  author={Butler, Daniel J and Wulff, Jonas and Stanley, Garrett B and Black, Michael J},
  booktitle={Computer Vision--ECCV 2012: 12th European Conference on Computer Vision, Florence, Italy, October 7-13, 2012, Proceedings, Part VI 12},
  pages={611--625},
  year={2012},
  organization={Springer}
}

@inproceedings{mehl2023spring,
  title={Spring: A high-resolution high-detail dataset and benchmark for scene flow, optical flow and stereo},
  author={Mehl, Lukas and Schmalfuss, Jenny and Jahedi, Azin and Nalivayko, Yaroslava and Bruhn, Andr{\'e}s},
  booktitle={Proceedings of the IEEE/CVF Conference on Computer Vision and Pattern Recognition},
  pages={4981--4991},
  year={2023}
}

@inproceedings{dosovitskiy2017carla,
  title={CARLA: An open urban driving simulator},
  author={Dosovitskiy, Alexey and Ros, German and Codevilla, Felipe and Lopez, Antonio and Koltun, Vladlen},
  booktitle={Conference on robot learning},
  pages={1--16},
  year={2017},
  organization={PMLR}
}

@inproceedings{ros2016synthia,
  title={The synthia dataset: A large collection of synthetic images for semantic segmentation of urban scenes},
  author={Ros, German and Sellart, Laura and Materzynska, Joanna and Vazquez, David and Lopez, Antonio M},
  booktitle={Proceedings of the IEEE conference on computer vision and pattern recognition},
  pages={3234--3243},
  year={2016}
}

@inproceedings{yu2020bdd100k,
  title={Bdd100k: A diverse driving dataset for heterogeneous multitask learning},
  author={Yu, Fisher and Chen, Haofeng and Wang, Xin and Xian, Wenqi and Chen, Yingying and Liu, Fangchen and Madhavan, Vashisht and Darrell, Trevor},
  booktitle={Proceedings of the IEEE/CVF conference on computer vision and pattern recognition},
  pages={2636--2645},
  year={2020}
}

@inproceedings{gaidon2016virtual,
  title={Virtual worlds as proxy for multi-object tracking analysis},
  author={Gaidon, Adrien and Wang, Qiao and Cabon, Yohann and Vig, Eleonora},
  booktitle={Proceedings of the IEEE conference on computer vision and pattern recognition},
  pages={4340--4349},
  year={2016}
}

@inproceedings{saunders2023self,
  title={Self-supervised Monocular Depth Estimation: Let's Talk About The Weather},
  author={Saunders, Kieran and Vogiatzis, George and Manso, Luis J},
  booktitle={Proceedings of the IEEE/CVF International Conference on Computer Vision},
  pages={8907--8917},
  year={2023}
}

@inproceedings{hurl2019precise,
  title={Precise synthetic image and lidar (presil) dataset for autonomous vehicle perception},
  author={Hurl, Braden and Czarnecki, Krzysztof and Waslander, Steven},
  booktitle={2019 IEEE Intelligent Vehicles Symposium (IV)},
  pages={2522--2529},
  year={2019},
  organization={IEEE}
}

@inproceedings{richter2017playing,
  title={Playing for benchmarks},
  author={Richter, Stephan R and Hayder, Zeeshan and Koltun, Vladlen},
  booktitle={Proceedings of the IEEE International Conference on Computer Vision},
  pages={2213--2222},
  year={2017}
}

@inproceedings{richter2016playing,
  title={Playing for data: Ground truth from computer games},
  author={Richter, Stephan R and Vineet, Vibhav and Roth, Stefan and Koltun, Vladlen},
  booktitle={Computer Vision--ECCV 2016: 14th European Conference, Amsterdam, The Netherlands, October 11-14, 2016, Proceedings, Part II 14},
  pages={102--118},
  year={2016},
  organization={Springer}
}

@misc{teed2020raft,
      title={RAFT: Recurrent All-Pairs Field Transforms for Optical Flow}, 
      author={Zachary Teed and Jia Deng},
      year={2020},
      eprint={2003.12039},
      archivePrefix={arXiv},
      primaryClass={cs.CV}
}

@InProceedings{flyingthings3d,
  author    = "N. Mayer and E. Ilg and P. H{\"a}usser and P. Fischer and D. Cremers and A. Dosovitskiy and T. Brox",
  title     = "A Large Dataset to Train Convolutional Networks for Disparity, Optical Flow, and Scene Flow Estimation",
  booktitle = "IEEE International Conference on Computer Vision and Pattern Recognition (CVPR)",
  year      = "2016",
  note      = "arXiv:1512.02134",
  url       = "http://lmb.informatik.uni-freiburg.de/Publications/2016/MIFDB16"
}

@misc{leastereo,
      title={Hierarchical Neural Architecture Search for Deep Stereo Matching}, 
      author={Xuelian Cheng and Yiran Zhong and Mehrtash Harandi and Yuchao Dai and Xiaojun Chang and Tom Drummond and Hongdong Li and Zongyuan Ge},
      year={2020},
      eprint={2010.13501},
      archivePrefix={arXiv},
      primaryClass={cs.CV}
}

@article{pandharipande2023sensing,
  title={Sensing and machine learning for automotive perception: A review},
  author={Pandharipande, Ashish and Cheng, Chih-Hong and Dauwels, Justin and Gurbuz, Sevgi Z and Ibanez-Guzman, Javier and Li, Guofa and Piazzoni, Andrea and Wang, Pu and Santra, Avik},
  journal={IEEE Sensors Journal},
  volume={23},
  number={11},
  pages={11097--11115},
  year={2023},
  publisher={IEEE}
}

@inproceedings{openpilot,
  title={Architecting Artificial Intelligence for Autonomous Cars: The OpenPilot Framework},
  author={Baresi, Luciano and Tamburri, Damian A},
  booktitle={European Conference on Software Architecture},
  pages={189--204},
  year={2023},
  organization={Springer}
}

@inproceedings{cui2024survey,
  title={A survey on multimodal large language models for autonomous driving},
  author={Cui, Can and Ma, Yunsheng and Cao, Xu and Ye, Wenqian and Zhou, Yang and Liang, Kaizhao and Chen, Jintai and Lu, Juanwu and Yang, Zichong and Liao, Kuei-Da and others},
  booktitle={Proceedings of the IEEE/CVF Winter Conference on Applications of Computer Vision},
  pages={958--979},
  year={2024}
}

@article{zhang2023perception,
  title={Perception and sensing for autonomous vehicles under adverse weather conditions: A survey},
  author={Zhang, Yuxiao and Carballo, Alexander and Yang, Hanting and Takeda, Kazuya},
  journal={ISPRS Journal of Photogrammetry and Remote Sensing},
  volume={196},
  pages={146--177},
  year={2023},
  publisher={Elsevier}
}

\end{document}